%% file: main.tex
\documentclass[letterpaper]{article}

\usepackage[utf8]{inputenc}
\usepackage{amsmath,mathtools,amssymb} % amssymb messes things with Bbbk
\usepackage[inline]{enumitem}
\usepackage[detect-weight=true, binary-units=true]{siunitx}
\usepackage{url,hyperref,cleveref}
\usepackage[linesnumbered]{algorithm2e}
\crefname{algocf}{algorithm}{algorithms}
\Crefname{algocf}{Algorithm}{Algorithms}
\usepackage{color,colortbl,etoolbox,xcolor} %http://tex.stackexchange.com/a/2768/22613
\robustify\bfseries
\usepackage{multirow,booktabs}
\usepackage{subcaption}
\usepackage{csquotes} %http://tex.stackexchange.com/a/294634
\usepackage[normalem]{ulem} %http://tex.stackexchange.com/a/23712
\renewcommand\vec{\boldsymbol}

\usepackage{setspace}
\usepackage{natbib,alifeconf}  %% The order is important

\input{plot-macros}

\title{Shape Change and Control of Pressure-based Soft Agents}
\author{Federico Pigozzi\\
\mbox{}\\
Evolutionary Robotics and Artificial Life lab, Department of Engineering and Architecture, University of Trieste\\
pigozzife@gmail.com}

\begin{document}
\maketitle

\begin{abstract}
% Abstract length should not exceed 250 words
  Biological agents possess bodies that are mostly of soft tissues.
  Researchers have resorted to soft bodies to investigate Artificial Life (ALife)-related questions; similarly, a new era of soft-bodied robots has just begun.
  Nevertheless, because of their infinite degrees of freedom, soft bodies pose unique challenges in terms of simulation, control, and optimization.
  Here we propose a novel soft-bodied agents formalism, namely Pressure-based Soft Agents (PSAs): they are bodies of gas enveloped by a chain of springs and masses, with pressure pushing on the masses from inside the body.
  Pressure endows the agents with structure, while springs and masses simulate softness and allow the agents to assume a large gamut of shapes.
  Actuation takes place by changing the length of springs or modulating global pressure.
  We optimize the controller of PSAs for a locomotion task on hilly terrain and an escape task from a cage; the latter is particularly suitable for soft-bodied agents, as it requires the agent to contort itself to squeeze through a small aperture.
  Our results suggest that PSAs are indeed effective at those tasks and that controlling pressure is fundamental for shape-changing.
  Looking forward, we envision PSAs to play a role in the modeling of soft-bodied agents, including soft robots and biological cells.\footnote{Videos of evolved agents are available at \url{https://pressuresoftagents.github.io}.}
\end{abstract}

\section{Introduction and related works}
\label{sec:introduction}
Softness is arguably one of the greatest gifts of mother nature.
Every living creature on Earth possesses a body that is mostly made of soft tissues.
Soft bodies can continuously bend, stretch, and twist, achieving adaptation to the environment; evolution keeps illuminating new ways to exploit softness, from the amazing manipulation feats of cephalopods \citep{hochner2012embodied}, to the protozoans of the genus \emph{Lacrymaria} \citep{mast1911habits}, that can contort their soft flagellum to grasp hard-to-reach preys, allowing for complex hunting dynamics to emerge.
It is not surprising that researchers have adopted soft materials to fabricate a new generation of soft robots \citep{rus2015design}, that promises to leverage shape change to recover from damages \citep{kriegman2019automated} and adapt to novel environments \citep{shah2021soft}.
In simulation, soft bodies are suitable to investigate virtual creatures for Artificial Life (ALife)-related questions \citep{joachimczak2016metamrphosis,kriegman2018morphological}, including evolutionary robotics \citep{cheney2014unshackling}.

At the same time, the simulation and optimization of soft agents pose unique challenges.
No analytical methods exist, as soft bodies have infinite degrees of freedom and entail, in general, hard-to-simulate dynamics \citep{laschi2016soft}.
Moreover, softness of bodies reinforces the paradigm known as \emph{embodied cognition} \citep{pfeifer2006body}, which posits a deep entanglement between the ``brain'' of an agent and the ``body'' that carries it \citep{pigozzi2022robots}.
While promising in terms of morphological computation \citep{nakajima2015information}, i.e., the brain offloading part of the computation to the body, such entanglement makes any co-optimization of brain and soft body arduous \citep{lipson2016difficulty}.
Finally, how to effectively achieve shape change remains an open issue in the literature \citep{shah2021shape}.

\begin{figure}
    \centering
    \subcaptionbox{\label{fig:psa}a PSA}{\includegraphics[width=0.45\linewidth,height=0.1\textheight]{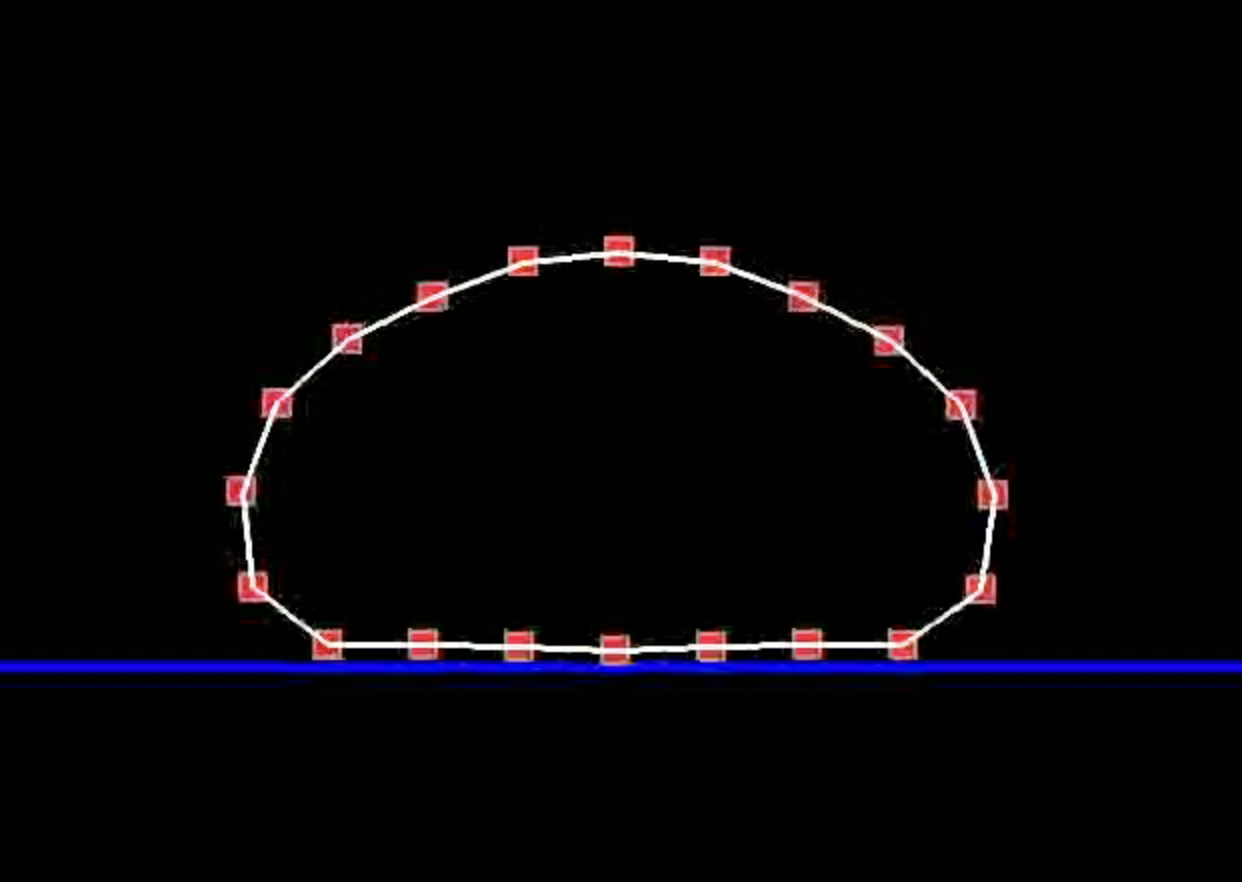}}
    \subcaptionbox{\label{fig:escaping}escaping from a cage}{\includegraphics[width=0.45\linewidth,height=0.1\textheight]{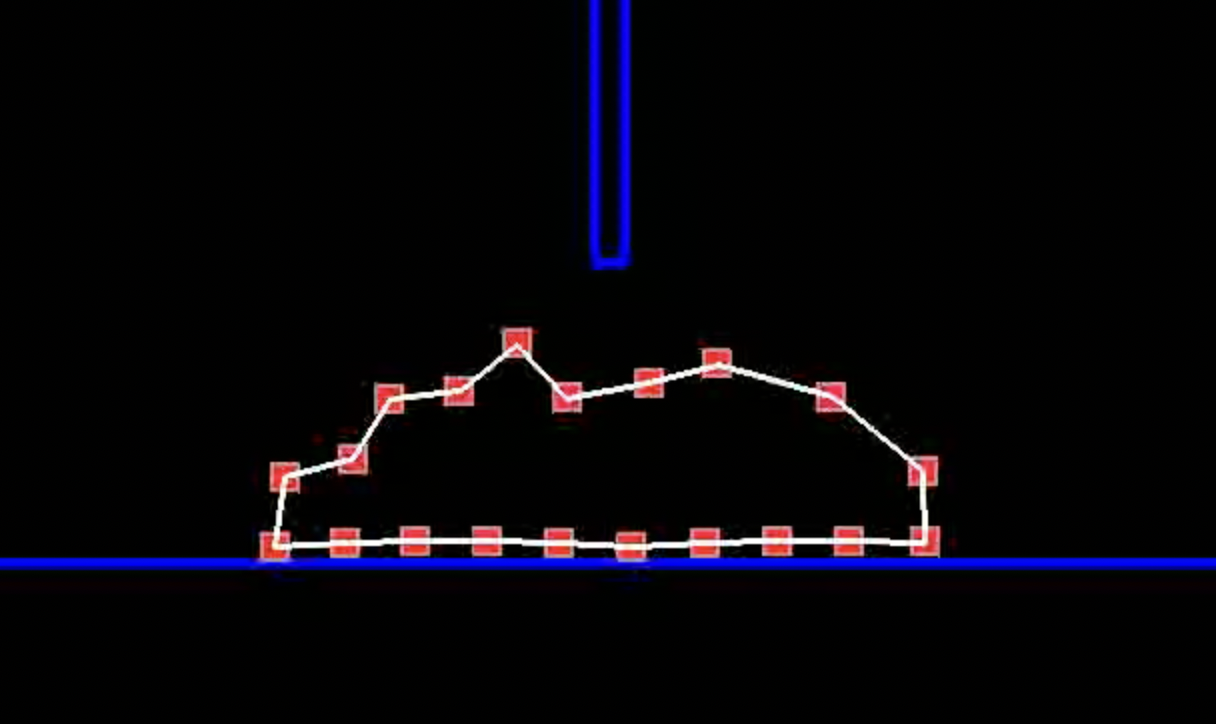}}
    \caption{
      (Left) Pressure-based Soft Agents (PSAs) are bodies of gas enveloped by a chain of springs and masses, with internal pressure endowing them with structure.
      Red squares are masses, white strings are springs, and blue shapes are environment bodies.
      (Right) PSAs can effectively achieve shape change to escape from a cage.
    }
    \label{fig:teaser}
\end{figure}

We propose a novel formalism to study soft-bodied agents, namely Pressure-based Soft Agents (PSAs).
They are bodies of gas enveloped by a chain of springs and masses, with pressure pushing on the masses from inside the body.
Pressure endows the agent with structure, while springs and masses simulate softness and allow the agent morphology to assume a large gamut of shapes, modelling the many degrees of freedom of soft bodies.
Actuation takes place by changing the resting length of springs and modulating global pressure.
We thoroughly describe the mechanical model and how to simulate it.
We also equip the agent with sensing abilities and a closed-loop controller that performs actuation by contracting or expanding the springs and changing global pressure.
See \Cref{fig:psa} for a snapshot of simulation.

Other soft agents formalisms do exist for virtual creatures, in particular, Voxel-based Soft Agents (VSAs) \citep{hiller2012automatic,medvet20202d,bhatia2021evolution}, which achieve softness by means of a spring-and-masses system, and Tensegrity-based Soft Agents (TSAs) \citep{rieffel2009automated,zappetti2017bio}, which achieve softness by connecting cables that are constantly in tension with rods that are constantly under compression.
Albeit far-reaching they might be, they still rely on an internal structure of rigid elements for the sake of modelling softness, severely limiting their ability to change shape.
Computer graphics, on the other side, employs also pressure-based soft bodies \citep{matyka2003pressure}, that rely on internal pressure to maintain structure and can thus stretch and bend in any possible configuration.
As a result, we ask ourselves whether it is possible to
\begin{enumerate*}[label=(\alph*)]
    \item \label{item:first} attain PSAs by endowing pressure-based soft bodies with a robotic controller, and
    \item \label{item:second} effectively exploit shape change for PSAs.
\end{enumerate*}

We experiment with a two-dimensional simulation of PSAs and carry out an extensive experimental campaign aimed at validating PSAs on two different tasks: a classic locomotion task on hilly terrain to answer \ref{item:first}, and an escape task from within a cage to answer \ref{item:second}.
The latter is particularly suitable to this work as it forces the agent to radically shape-shift in order to escape through an aperture in the cage.
We experiment with PSAs of three different sizes and optimize their controller with an established numerical optimizer \citep{hansen2001cmaes}.

Our results suggest that PSAs are indeed proficient at solving both traditional tasks---locomotion---and tasks that require changing shape---escape.
Moreover, we also show that preventing the controller from modulating pressure (i.e., pressure is the result of only physical interactions) makes it impossible for PSAs to solve the tasks.

Looking forward, we believe PSAs can play a role in the simulation of soft-bodied agents.
Indeed, many existing soft robots rely on pressure to shape change, by means of pumps \citep{kriegman2021scale,shah2021soft} or inflatable tubes \citep{usevitch2020untethered,drotman2021electronics}.
Finally, we envision many exciting ALife applications, including the modelling of biological cells that, similarly to PSAs, consist in a fluid, the cytoplasm, enveloped by a flexible membrane.

\section{Proposed agent model}
\label{sec:model}
We propose a simple, yet expressive model of soft agents, namely Pressure-based Soft Agents (PSAs).
They are bodies of gas contained within an envelope (a chain) of springs and masses, with pressure pushing on the masses from inside the body: actuation takes place by
\begin{enumerate*}[label=(\alph*)]
    \item contracting or expanding the springs, and
    \item changing pressure.
\end{enumerate*}
The harmonious execution of these two allows the PSA to assume a large gamut of shapes.
By virtue of their many degrees of freedom, PSAs are both
\begin{enumerate*}
    \item expressive, and
    \item challenging to control.
\end{enumerate*}

We take inspiration from the work of \citet{matyka2003pressure} on pressure-based soft bodies for computer graphics.
Such model is particularly suitable for bodies that can bend and twist in arbitrary shapes, as balloons and cloth; as a result, we introduce it to soft agents.
To ease modelling, we work with a two-dimensional simulation in discrete time and continuous space.
However, we remark that the representations and algorithms of this work are easily portable to the three-dimensional setting.

We define a PSA as the combination of an embodiment, which obeys a \emph{mechanical model} and possesses \emph{sensing} capacities, and a brain, which we implement with a \emph{controller}.

\subsection{Mechanical model}
\label{sec:mechanical}

\subsubsection{Morphology}
\label{sec:morphology}
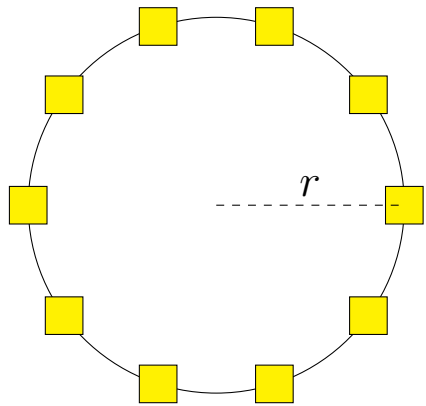
\begin{figure}
    \centering
    \begin{tikzpicture}
        \def \n {10}
        \def \radius {2.5}
        \draw circle(\radius)
              foreach \s in{1,...,\n}{
                  (-360/\n*\s:-\radius)
                  node[anchor=center,rectangle,draw,fill=yellow,minimum width=0.5cm,minimum height=0.5cm] at (-360/\n*\s:-\radius) {}
              };
        \draw [dashed] (0,0) -- (\radius,0);
        \draw node at (\radius/2,0.25) {\huge $r$};
    \end{tikzpicture}
    \caption{
      The building blocks of a PSA morphology of radius $r$: yellow squares are masses, black strings are springs.
    }
    \label{fig:morphology}
\end{figure}

A PSA morphology is a \emph{body} of gas contained within an \emph{envelope}.
We define an envelope from a circle of radius $r$; for simplicity, let us assume its center is the origin.
We place $n_\text{mass}$ masses of rigid material equispaced along the circumference, i.e., at points $r + \cos \frac{2\pi i}{n_\text{mass}}, r + \sin \frac{2\pi i}{n_\text{mass}}$, and fix their rotation.
We join each mass with the previous and the next masses along the circumference with distance joints of frequency $f$, damping ratio $d$, maximum length $l_\text{max}$, minimum length $l_\text{min}$.
Moreover, there exists an internal pressure $p$ (in \si{\pascal}) that acts on the masses; with no pressure, the envelope would collapse because of gravity.
Pressure thus endows the body with structure.
We remark $p$ is global, in the sense that it is the same for all masses.
We summarize the building blocks of a PSA morphology in \Cref{fig:morphology}.

The masses define the boundaries of the morphology and collide with external bodies.
The joints, by choosing appropriate values for $f$ and $d$, act as springs: they contract and expand in response to forces acting on the masses they join.
As a result, the envelope is not rigid but soft, and the morphology deforms under forces acting on the masses, either exogenous, e.g., contact with other bodies, or endogenous, i.e., changes in $p$.

We remark that, indeed, spring-and-damper systems are at the heart of other soft agents simulators, including VSAs \citep{hiller2012automatic,medvet20202d} and TSAs \citep{zappetti2017bio}.
Moreover, springs allow masses to change their relative position, endowing the mechanical model with many degrees of freedom; in fact, the envelope can stretch and bend, and the body can contract or expand in limitless configurations.
By virtue of such freedom, our model is suitable for modelling the infinite degrees of freedom of soft bodies, including soft robots.
At the same time, such freedom entails that computing the area of the morphology is not tractable: with PSAs, we solve this problem by indirectly updating the area with $p$, in a way that we detail in the next paragraph.

As an aside, PSAs can be seen not only as robotic agents, but also as a minimal model of a cell: the envelope constitutes the cellular membrane \citep{singleton2004bacteria}, with masses playing the role of membrane proteins and springs the role of lipids.
Being fluid, the gas effectively models the cytoplasm \citep{shepherd2006cytomatrix}.
Finally, $p$ closely resembles turgor pressure acting on the membrane \citep{pritchard2001turgor}.

\subsubsection{Simulation}
\label{sec:simulation}
Area $a$ (in \si{\meter^2}) alters according to pressure $p$; $p$, in turn, can be the output of a controller or change according to physical laws.
Since pressure is what endows PSAs with structure, we treat the latter as an ablation study in the Results section, and focus on PSAs that control $\Delta p$ (thus affecting $p$).

At every time step of simulation, we compute the total pressure acting on the side of a joint, and distribute it over the masses.
In detail, we:

\begin{enumerate}[label=(\arabic*)]
    \item query the controller for $\Delta p$, and sum it to $p$.
    \item for every $i$-th joint, compute the total pressure acting on its side as $p_i=l_i p$, where $l_i$ is the length of the joint, and the normalized normal vector $\hat{\vec{n}}_i \in [-1,1]^2$ pointing to the interior of the morphology.
    \item for every $j$-th mass, let $i^-$ and $i^+$ be the joints joining it to the previous and next masses in the envelope, respectively.
    We transform scalar pressure into directed pressure forces $\vec{p}_{j,i^-}=\frac{p_{i^-}}{2} \hat{\vec{n}}_{i^-}$ and $\vec{p}_{j,i^+}=\frac{p_{i^+}}{2} \hat{\vec{n}}_{i^+}$ acting on the two joints.
    We divide by $2$ to equally distribute pressure on the the masses that anchor a joint.
    \item for every $j$-th mass, we compute  $\vec{p}_j = \vec{p}_{j,i^-} + \vec{p}_{j,i^+}$ and apply it as a force to the mass center.
    We remark that $\vec{p}_j$ is indeed in \si{\newton}, as $\vec{\hat{n}}$ is dimensionless, $l_i$ is in \si{\meter}, and $p$ is in \si{\pascal}, with $\SI{1}{\pascal}=\SI{1}{\newton \meter^{-1}}$.
    \item step the physics engine.
\end{enumerate}

Thus, $a$ is not a free parameter (as $p$), but we affect it through pressure, as higher pressure on the masses implies larger area, and vice versa.
Finally, the overall shape of the PSA morphology, i.e., the arrangement and relative positions of the masses, depends on contacts with other bodies, and changes in the resting length of springs dictated by the controller.

\subsubsection{Parameters}
\label{sec:parameters}
Masses are squares of side \SI{1}{\meter} and density \SI{2500}{\kilogram\ \meter^{-2}}; we found results to be consistent also with other sizes and densities.
After preliminary experiments and relying on our previous knowledge, we set $f=\SI{8}{\hertz}$, $d=0.3$, $l_\text{max}=1.25l$, and $l_\text{min}=0.75l$, where $l$ is the resting length of a spring.
As far as $r$ and $n_\text{mass}$ are concerned, they vary according to the morphology to simulate, as we shall see in the next section.

\subsection{Sensing}
\label{sec:sensing}
In order to implement a closed-loop controller, we equip PSAs with sensors.
Indeed, sensing is an important property for agents that interact with an environment \citep{talamini2019evolutionary}.
In this work, we employ touch, pressure, position, and velocity sensors.
Touch sensors perceive whether masses are touching other bodies (e.g., the ground) or not, and, for each mass, return $1$ if yes, $0$ otherwise.
The pressure sensor perceives the current internal pressure $p$, and is thus a proprioceptor.
Position sensors perceive the relative $x$- and $y$-position of each mass from the center of mass of the morphology.
Finally, velocity sensors perceive the $x$- and $y$-velocity of the center of mass of the body.

We normalize every sensor reading into $[0,1]$, and, to introduce sensory memory, compute its average over the last $t$ time steps (using the normalized values).
We then concatenate all the sensor readings into an observation vector $\vec{o} \in [0,1]^{3n_\text{mass} + 3}$.
After preliminary experiments, we set $t=25$.

\subsection{Controller}
\label{sec:controller}
At every time step of simulation, we feed the current observation vector $\vec{o}$ to a controller that decides two sets of actions:
\begin{enumerate*}[label=(\alph*)]
    \item the resting length of the springs, and
    \item the change in pressure $\Delta p$.
\end{enumerate*}

As far as the former is concerned, given a control value $s \in [-1,1]$, we instantaneously modify the resting length $l$ of a spring as:
\begin{equation}
    \label{eq:actuation}
    l =
    \begin{cases}
        l - s(l - l_\text{min}) &\text{if}\ s > 0\\
        l &\text{if}\ s=0\\
        l - s(l_\text{max} - l) &\text{if}\ s < 0\\
    \end{cases}
\end{equation}
Thus, $s=-1$ corresponds to the maximum expansion, and $s=1$ corresponds to the maximum contraction, all other values lying in between.
This is the same model of actuation of other soft robotics simulators, e.g., \citep{medvet20202d,bhatia2021evolution}.

Change in pressure, on the other side, requires a different domain, as $[-1,1]$ is not morphology-agnostic, given that different morphologies require different pressure ranges.
As a result, we split the controller into a \emph{pressure controller} $\pi_p$ and a \emph{springs controller} $\pi_s$.
The former takes as input $\vec{o}$ and outputs $\Delta p$, that we clip to $[p_\text{min},p_\text{max}]$ in order to remain within meaningful boundaries; the latter takes as input $\vec{o}$ and outputs $\vec{s} \in [-1,1]^{n_\text{mass}+1}$ (i.e., one control value for every spring).

After preliminary experiments, we implemented $\pi_p$ as a linear model of the form:
\begin{equation}
    \Delta p = \vec{W}_p \vec{o} + b_p
    \label{eq:linear}
\end{equation}
with weights $\vec{W}_p \in \mathbb{R}^{1 \times |\vec{o}|}$ and bias $b_p \in \mathbb{R}$.
As a result, $\Delta p \in \mathbb{R}$ and the model can choose the output most appropriate to its morphology.
Similarly, we implemented $\pi_s$ with a non-linearity to ensure the output lies in $[-1,1]$:
\begin{equation}
    \vec{s} = \tanh(\vec{W}_s \vec{o} + \vec{b}_s)
    \label{eq:tanh}
\end{equation}
with weight matrix $\vec{W}_s \in \mathbb{R}^{(n_\text{mass} + 1) \times |\vec{o}|}$ and bias vector $\vec{b}_s \in \mathbb{R}^{n_\text{mass} + 1}$.

We focus on optimizing the controller of a PSA for a task. Thus, the parameters we optimize are the controller parameters $\vec{\theta} = [\vec{\theta}_p \; \vec{\theta}_s]$, where $\vec{\theta}_p=[\vec{W}_p \; b_p]$ are the pressure controller parameters, and $\vec{\theta}_s=[\vec{W}_s \; \vec{b}_s]$ are the springs controller parameters.

\section{Experimental procedure}
\label{sec:procedure}
We performed an experimental campaign aimed at answering the following research questions:

\begin{enumerate}[label=RQ\arabic*,leftmargin=2.75\parindent]
  \item \label{item:rq-0} Is the mechanical model valid to simulate pressure-based soft bodies?
  \item \label{item:rq-1} Can we control PSAs? In other words, are PSAs capable of solving a classic locomotion task on hilly terrain?
  \item \label{item:rq-2} Can we effectively exploit shape change for PSAs?
\end{enumerate}

We design a specific task for each question; we detail the tasks in the next sub-section.

For all tasks, we evaluate three different PSA morphologies, in order to get a sense of the effectiveness of PSAs across a wide array of morphological conditions.
For the \emph{large} morphology, we set $n_\text{mass}=20$ and $r=\SI{10}{\meter}$; for the \emph{medium} morphology, we set $n_\text{mass}=15$ and $r=\SI{7.5}{\meter}$; finally, for the \emph{small} morphology, we set $n_\text{mass}=10$ and $r=\SI{5}{\meter}$.
For the three morphologies, the input size is $33$, $48$, and $63$, respectively.
As a result, the size of the parameter space $|\vec{\theta}|$ is $408$, $833$, and $1408$, respectively.

Since pressure is what endows PSAs with structure, we investigate whether it is really necessary or not to accomplish the tasks and conduct the following ablation study in \ref{item:rq-1} and \ref{item:rq-2}.

\subsection{Ablation}
\label{sec:ablation}
As an ablation study, we experiment with a configuration \emph{without pressure control}, in contrast to the configuration \emph{with pressure control} considered so far.
To this end, we dispense with the pressure controller $\pi_p$ and let $p$ be the result of physical laws.
The pressure of an ideal gas changes according to the ideal gas law of \citet{clapeyron1834memoire}:

\begin{equation}
    pa = nRT
    \label{eq:ideal}
\end{equation}
where $p$ (in \si{\pascal}) is the pressure value, $n$ is the amount of substance (in \si{\mol}), $R$ is the ideal gas constant (in \si{\meter^2\pascal\ \mol^{-1}\kelvin^{-1}}), and $T$ is temperature (in \si{\kelvin}).
We remark that many real gases do behave as ideal under various temperature and pressure conditions \citep{cengel2011thermodynamics}.
By fixing $T$, the right-hand side of \Cref{eq:ideal} is constant: then, $p$ must change to accommodate changes in $a$ and balance the equation.
At every time step, we compute $a$ by triangulation and plug it into \Cref{eq:ideal} to compute $p$.

We set $T=\SI{288.15}{\kelvin}=\SI{15}{\celsius}$ to simulate room temperature, and the gas (the PSA is filled with) to be $\mathrm{N_2}$ (nitrogen), a cheap and common gas.
$n$ is the ratio between the gas mass $m$ (in \si{\kilogram}) and the molar mass (in \si{\kilogram\ \mol^{-1}}), that is $\SI{0.0280314}{\kilogram\ \mol^{-1}}$ for $\mathrm{N_2}$.
We set $m=\SI{0.1}{\kilogram}$, $m=\SI{0.075}{\kilogram}$, and $m=\SI{0.05}{\kilogram}$ for the three morphologies respectively.
As usual, $R=\SI{8.3145626}{\meter^2\pascal\ \mol^{-1}\kelvin^{-1}}$ is the ideal gas constant.

For this configuration, the size of the parameter space $|\vec{\theta}|$ is $374$, $784$, and $1334$, respectively for the three morphologies; thus, disabling pressure control results not only in simpler actuation, but also in a smaller search space that might benefit optimization.

Finally, for the configuration with pressure control, we set $p_\text{max}=\frac{nRT}{\pi r^2}$, where $\pi r^2$ is the area of a perfect circle of radius $r$ and $p_\text{min}=0.2p_\text{max}$ to prevent the PSA from collapsing.

\subsection{Tasks}
\label{sec:tasks}
\begin{figure}
    \centering
    \subcaptionbox{\label{fig:locomotion}locomotion}{\includegraphics[width=0.45\linewidth,height=0.1\textheight]{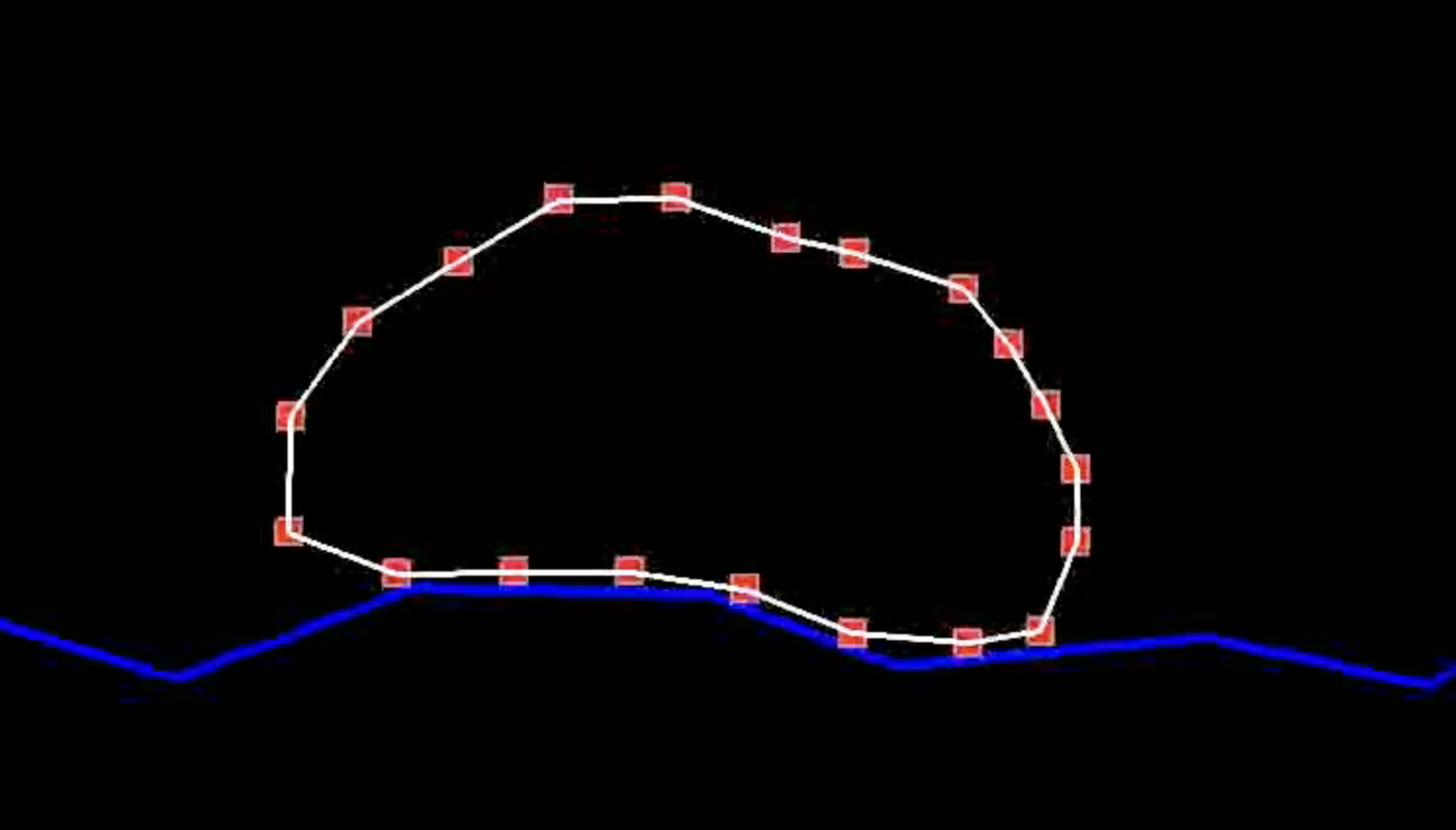}}
    \subcaptionbox{\label{fig:escape}escape}{\includegraphics[width=0.45\linewidth,height=0.1\textheight]{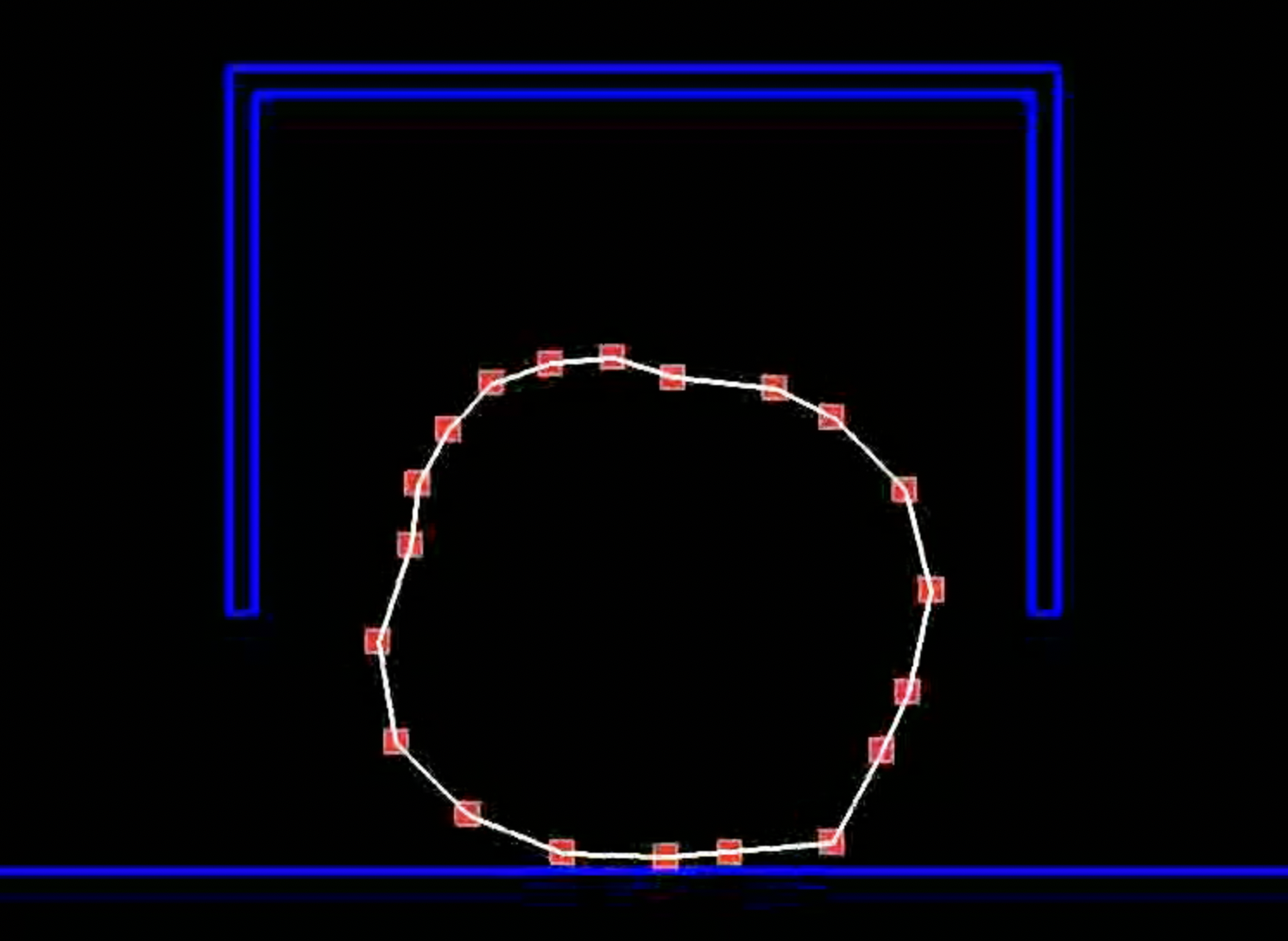}}
    \caption{
      The tasks considered in our experiments.
    }
    \label{fig:tasks}
\end{figure}

We evaluate our method on two tasks: locomotion and escape.
See \Cref{fig:tasks} for sample frames from these tasks.

\subsubsection{Locomotion}
\label{sec:locomotion}
Locomotion is a classic task in evolutionary robotics \citep{sims1994evolving,nolfi2000evolutionary}, and provides a benchmark of basic control skills.
It consists in walking as fast as possible over a terrain along the $x$ direction, over an amount of simulated time $t_\text{final}$.
The fitness function is the velocity $\overline{v}_x$ of the center of mass of the PSA over the simulation.
We set $t_\text{final}=\SI{30}{\second}$.
While the terrain is usually a flat surface, we here consider a more challenging hilly terrain, with bumps of different heights and distances.
For a given seed, we randomly procedurally generate the bumps with an average height of \SI{1}{\meter}, \SI{2}{\meter}, and \SI{3}{\meter} for the three morphology sizes, and an average distance of \SI{10}{\meter}.

\subsubsection{Escape}
\label{sec:escape}
Escape is particularly suitable for soft agents \citep{cheney2015evolving}, as it forces the agent to radically change its shape to pass through an aperture.
At the onset of each simulation, we place the PSA within a cage.
The cage amounts to a roof and two walls, with one small aperture per side.
The task consists in escaping as fast as possible in any direction over a maximum amount of simulated time $t_\text{final}$.
The fitness function is the average velocity $\overline{v}$ of the center of mass of the PSA over the simulation, regardless of the direction.
We set $t_\text{final}=\SI{30}{\second}$.
The cage is rigid, immobile, and indestructible, forcing the PSA to contort itself and squeeze through one of the apertures.
After preliminary experiments, for a PSA of radius $r$, we set the roof height to $2r+1$, the walls $3r$ apart from each other, and the apertures one third of the roof height.
Escape differs from locomotion in that there is a clear-cut condition for ``solving'' it, namely when all of the PSA masses are outside of the cage: if this is the case, we terminate the simulation.

\subsection{Optimization}
\label{sec:optimization}
We optimize the controller parameters $\vec{\theta}$ with Covariance Matrix Adaptation Evolution Strategy (CMA-ES) \citep{hansen2001cmaes,hansen2016cma}, an established numerical optimizer.
While it is possible to use any optimization algorithm, we found CMA-ES to be stable across the different tasks, also thanks to the small size of the search space \citep{muller2018challenges}.
CMA-ES iteratively optimizes the solution in the form of a multivariate normal distribution, against a given fitness function.
At each iteration, it samples the distribution obtaining a population of solutions and then updates the parameters of the distribution based on the best half of the population.
CMA-ES employs non-trivial heuristics while updating the distribution---we refer the reader to \citet{hansen2001cmaes} for more details.

We use the default parameters suggested in \citep{hansen2006cma}, namely the initial step size $\sigma=0.5$ and the population size $\lambda=3 \lfloor \log |\vec{\theta}| \rfloor$.
We set the initial vector of means by sampling uniformly the interval $[-1,1]$ for each vector element.
We let CMA-ES iterate until \num{10000} fitness evaluations have been done.

\subsection{Settings}
For each experiment, we performed $5$ evolutionary runs by varying the random seed for CMA-ES and the terrain generation in locomotion.
We carried out all statistical tests with the Mann-Whitney U rank test for independent samples.
We employ as physics engine the \text{Python} wrapper\footnote{\url{https://github.com/pybox2d/pybox2d}} to \text{Box2D} \citep{catto2011box2d}, a popular 2D physics library written in \text{C++}.
We set the simulation frequency to \SI{60}{\hertz} and left all other parameters unchanged.
We remark that, for a given seed and controller, all simulations are deterministic.
For CMA-ES, we used the implementation of \citet{ha2017evolving}, that is a wrapper around the \text{pycma} library \citep{hansen2019cma}, and, at a given iteration, parallelize fitness evaluations using multiprocessing.
Each run took approximately \SI{1}{\hour} on an Apple M1 MacBook Pro at \SI{3.2}{\giga\hertz} with \SI{8}{\giga\byte} \text{RAM} and $8$ cores.
We made the code publicly available at \url{https://github.com/pigozzif/PressureSoftAgents}.

\section{Results}
\label{sec:results}

\subsection{\ref{item:rq-0}: validation of the mechanical model}
\label{sec:rq0}
We validate whether the proposed mechanical model is suitable for simulating pressure-based soft bodies.
In particular, for a PSA of radius $r$, we verify if there exists a $p$ such that $a$ is that of a perfect circle of radius $r$; in this way, we assess whether our model can correctly simulate a balloon---an ideal pressure-based soft body.

To this end, we conduct the following experiment:
\begin{enumerate}[label=(\alph*)]
    \item We define a controller that, for a morphology of $n_\text{mass}$ masses and radius $r$, outputs $\vec{s} \in \vec{0}^{n_\text{mass} + 1}$ and $\Delta p = \frac{p_\text{max}}{100}$ at every time step.
    In other words, it does not alter the resting length of springs, while constantly increasing the pressure.
    \item For each of the morphologies, we run a simulation on flat terrain using the aforementioned controller, setting $p=p_\text{min}=0$ at the beginning.
    \item For each time step of simulation, we record pressure $p$ and $\rho=\frac{a}{\pi r^2}$ as performance indexes, $\pi r^2$ being the area of a perfect circle of radius $r$.
\end{enumerate}
If our proposed mechanical model correctly simulates pressure-based soft bodies, there must exist a value $p$ such that $\rho=1$.

\begin{figure}%[ht!]
    \centering
    \begin{tikzpicture}
        \begin{groupplot}[
            width=\linewidth,
            height=0.5\linewidth,
            group style={
                group size=1 by 1,
                horizontal sep=1.5mm,
                vertical sep=1.5mm,
                xticklabels at=edge bottom,
                yticklabels at=edge left
            },
            every axis plot/.append style={thick},
            scaled x ticks = false,
            grid=both,
            grid style={line width=.1pt, draw=gray!10},
            major grid style={line width=.15pt, draw=gray!50},
            minor tick num=5,
            xlabel={Time steps},
            ymin=0,ymax=1.2
        ]
            \nextgroupplot[
                align=center,
                legend columns=4,
                legend entries={Large,Medium,Small,Pressure},
                legend to name=legendRQ0,
                ylabel={$\rho,\frac{p}{p_\text{max}}$}
            ]
            \linesimple{data/line/val.txt}{t}{large}{cola1}
            \linesimple{data/line/val.txt}{t}{medium}{cola2}
            \linesimple{data/line/val.txt}{t}{small}{cola3}
            \linesimple{data/line/val.txt}{t}{p}{cola4}
            
        \end{groupplot}
    \end{tikzpicture}
    \pgfplotslegendfromname{legendRQ0}
    \caption{
        Ratio $\rho$ between the PSA area $a$ and the area of a perfect circle of the same radius, together with relative pressure $\frac{p}{p_\text{max}}$, obtained with three sizes.
        Our proposed mechanical model effectively simulates pressure-based soft bodies, as $\rho$ approaches $1$ by constantly increasing pressure.
    }
    \label{fig:rq0}
\end{figure}
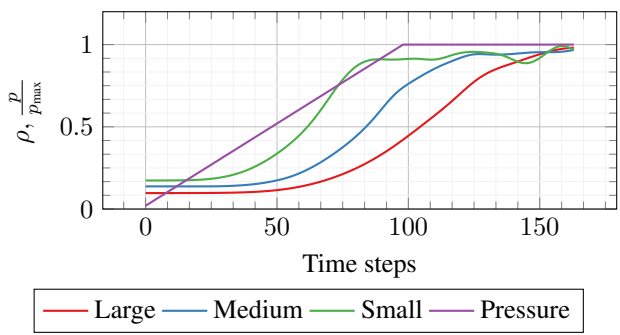

We report the results in \Cref{fig:rq0}.
According to the figure, the results are qualitatively similar for the three morphologies.
Area starts off just above $0$, since $p=0$ and there is no pressure supporting the envelope; it then smoothly increases throughout the simulation, before plateauing at $1$ after $p=p_\text{max}$.

Through that evidence, we can answer positively to \ref{item:rq-0}: our proposed mechanical model is suitable for simulating pressure-based soft bodies.

\subsection{\ref{item:rq-1}: can we control PSAs?}
\label{sec:rq1}
In order to validate the effectiveness of our proposed agent model, we measure the performance of PSAs in a classic locomotion task, in two different settings: with and without pressure control.
In both cases, we use $\overline{v}_x$ as performance index.

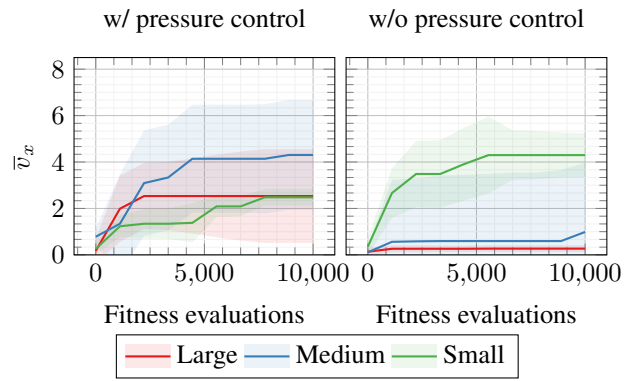
\begin{figure}%[ht!]
    \centering
    \begin{tikzpicture}
        \begin{groupplot}[
            width=0.6\linewidth,
            height=0.5\linewidth,
            group style={
                group size=2 by 1,
                horizontal sep=1.5mm,
                vertical sep=1.5mm,
                xticklabels at=edge bottom,
                yticklabels at=edge left
            },
            every axis plot/.append style={thick},
            scaled x ticks = false,
            grid=both,
            grid style={line width=.1pt, draw=gray!10},
            major grid style={line width=.15pt, draw=gray!50},
            minor tick num=5,
            xlabel={Fitness evaluations},
            ymin=0,ymax=8.5
        ]
            \nextgroupplot[
                align=center,
                legend columns=3,
                legend entries={Large,Medium,Small},
                legend to name=legendRQ1evo,
                ylabel={$\overline{v}_x$},
                title={w/ pressure control}
            ]
            \linewitherror{data/line/evo.hilly.txt}{i}{large_mu}{large_std}{cola1}
            \linewitherror{data/line/evo.hilly.txt}{i}{medium_mu}{medium_std}{cola2}
            \linewitherror{data/line/evo.hilly.txt}{i}{small_mu}{small_std}{cola3}
            
            \nextgroupplot[
                title={w/o pressure control}
            ]
            \linewitherror{data/line/evo.hilly.ablation.txt}{i}{large_mu}{large_std}{cola1}
            \linewitherror{data/line/evo.hilly.ablation.txt}{i}{medium_mu}{medium_std}{cola2}
            \linewitherror{data/line/evo.hilly.ablation.txt}{i}{small_mu}{small_std}{cola3}
        \end{groupplot}
    \end{tikzpicture}
    \pgfplotslegendfromname{legendRQ1evo}
    \caption{
        Median $\pm$ standard deviation (solid line and shaded area) of the average velocity of locomotion for the best individuals found during each evolutionary run, obtained with three sizes and with or without pressure control.
        Our agent model is effective at locomotion on hilly terrain.
    }
    \label{fig:rq1-evo}
\end{figure}

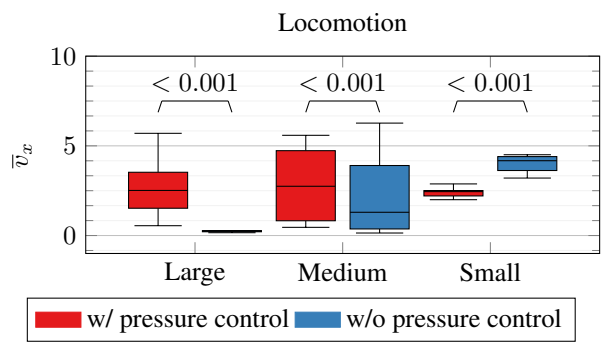
\begin{figure}
    \centering
    \begin{tikzpicture}
        \begin{groupplot}[
            boxplot,
            boxplot/draw direction=y,
            width=\linewidth,
            height=0.5\linewidth,
            group style={
                group size=1 by 1,
                horizontal sep=1.5mm,
                vertical sep=1.5mm,
                yticklabels at=edge left
            },
            ymin=-1,ymax=10,
            legend cell align={left},
            ymajorgrids=true,
            yminorgrids=true,
            grid style={line width=.1pt, draw=gray!10},
            major grid style={line width=.15pt, draw=gray!50},
            minor y tick num=5,
            xmajorticks=true,
            xminorticks=false,
            xticklabels={{Large}, {Medium}, {Small}},
            xtick={1.5,3.5,5.5},
            title={Locomotion}
        ]
            \nextgroupplot[
                align=center,
                legend columns=2,
                legend entries={w/ pressure control,w/o pressure control},
                ylabel={$\overline{v}_x$},
                legend to name=legendRQ1box
            ]
            \addlegendimage{area legend,color=cola1,fill}
            \addlegendimage{area legend,color=cola2,fill}
            \addplot[black,fill=cola1] table[y=large] {data/box/box.hilly.txt};
            \addplot[black,fill=cola2] table[y=large_ablation] {data/box/box.hilly.txt};
            \addplot[black,fill=cola1] table[y=medium] {data/box/box.hilly.txt};
            \addplot[black,fill=cola2] table[y=medium_ablation] {data/box/box.hilly.txt};
            \addplot[black,fill=cola1] table[y=small] {data/box/box.hilly.txt};
            \addplot[black,fill=cola2] table[y=small_ablation] {data/box/box.hilly.txt};
            \pvalue{1}{2}{7.5}{$<0.001$}
            \pvalue{3}{4}{7.5}{$<0.001$}
            \pvalue{5}{6}{7.5}{$<0.001$}
            
        \end{groupplot}
    \end{tikzpicture}
    \\
    \pgfplotslegendfromname{legendRQ1box}
    \caption{
        Distribution of the average velocity of locomotion for the best individuals found for each evolutionary run, obtained with three sizes and with or without pressure control.
        Dispensing with pressure control generally hampers performance in locomotion on hilly terrain.
        Numbers are $p$-values.
    }
    \label{fig:rq1-box}
\end{figure}

We summarize the results in \Cref{fig:rq1-evo}, which plots $\overline{v}_x$ in terms of median $\pm$ standard deviation for the best individuals over the course of evolution.
Moreover, \Cref{fig:rq1-box} reports boxplots for the distribution of $\overline{v}_x$ of the best individuals.
For every morphology, we also show the $p$-value for the statistical test against the null hypothesis of equality between the medians with and without pressure control.

From the figures, we see that our proposed agent model is effective at the task of locomotion and succeeds in mastering it, regardless of the morphology.
We visually inspected the behaviors and found them to be highly adapted for a locomotion task on hilly terrain.
PSAs evolve to ``roll'' over the ground, sliding the masses one after the other, and modulating pressure in order to have the right shape to overcome bumps: in fact, we found that decreasing pressure right before a bump allows the PSA to lower its center of gravity and generate enough momentum to walk over it.
On the other side, increasing pressure on flat portions of terrain allows the PSA to bounce over it and generate enough momentum to walk faster.
Interestingly, we found some individuals to show life-like behaviors: as a matter of example, when approaching bumps, some stretched out their front (masses and joints) to reach over the tip of the bump, grasp it, and finally walk over it.
Others appeared to adopt the same strategy to ``probe'' the terrain in front of them, and plan future actions accordingly.
We made videos with pressure control available at \url{https://pressuresoftagents.github.io}.

According to the figures, PSAs evolved without pressure control were not as effective.
In two morphologies out of three, $\overline{v}_x$ does not even depart from its initial value, meaning that no adaptation takes place; surprisingly, the same is not true for the Small morphology, which even succeeds in outperforming its counterpart with pressure control.
We remark that, as shown in \Cref{fig:rq1-box}, $p$-values are significant for all three comparisons.
To gain further insights into this phenomenon, we visually inspected the evolved behaviors without pressure control.
We found them to be not adapted to a locomotion task on hilly terrain.
In particular, Medium and Large PSAs often get stuck in hollows of the terrain; other times, they unsuccessfully struggle to walk over a bump.
We believe the reason to be the lack of pressure control: as mentioned before, modulating pressure allows the PSA to deform according to the terrain at hand.
This fact also hints at why Small PSAs are effective even without pressure control: thanks to their small size, contacts with the terrain body are relatively enough to allow for sufficient deformation.

%As a side comment, we also experimented with phase controllers, as they are very common in robotics for locomotion tasks, also for soft robots \citep{cheney2014unshackling}.
%In synthesis, a phase controller uses as actuation the output of a sinusoidal function of a frequency, an amplitude, and phase parameters that we optimize.
%We found phase controllers to deliver abysmal performance, and do not report the results for the sake of brevity.
%The reason is likely to be that effective behavior for PSAs requires rolling over the terrain, and so must exploit sensing---that phase controllers do not have---to perceive the current geometry of the morphology.

Through that evidence, we can answer positively to \ref{item:rq-1}: we can conclude that, after optimization, it is possible to control PSAs for a task requiring a decent level of cognition, considering the challenging nature of the hilly terrain.
Moreover, ablating the pressure control component of the controller results in much worse performance, especially for bigger (and, we believe, more realistic) morphologies, suggesting that pressure control is an inextricable part of our proposed agent model.

\subsection{\ref{item:rq-2}: can we effectively exploit shape change?}
\label{sec:rq2}
In order to assess the shape-changing abilities of our proposed agent model, we measure the performance of PSAs in an escape task, in two different settings: with and without pressure control.
In both cases, we use $\overline{v}$ as performance index.

\begin{figure}
    \centering
    \begin{tikzpicture}
        \begin{groupplot}[
            boxplot,
            boxplot/draw direction=y,
            width=\linewidth,
            height=0.5\linewidth,
            group style={
                group size=1 by 1,
                horizontal sep=1.5mm,
                vertical sep=1.5mm,
                yticklabels at=edge left
            },
            ymin=0,ymax=1.5,
            legend cell align={left},
            ymajorgrids=true,
            yminorgrids=true,
            grid style={line width=.1pt, draw=gray!10},
            major grid style={line width=.15pt, draw=gray!50},
            minor y tick num=5,
            xmajorticks=true,
            xminorticks=false,
            xticklabels={{Large}, {Medium}, {Small}},
            xtick={1.5,3.5,5.5},
            title={Escape}
        ]
            \nextgroupplot[
                align=center,
                legend columns=2,
                legend entries={w/ pressure control,w/o pressure control},
                ylabel={$\overline{v}$},
                legend to name=legendRQ2box
            ]
            \addlegendimage{area legend,color=cola1,fill}
            \addlegendimage{area legend,color=cola2,fill}
            \addplot[black,fill=cola1] table[y=large] {data/box/box.escape.txt};
            \addplot[black,fill=cola2] table[y=large_ablation] {data/box/box.escape.txt};
            \addplot[black,fill=cola1] table[y=medium] {data/box/box.escape.txt};
            \addplot[black,fill=cola2] table[y=medium_ablation] {data/box/box.escape.txt};
            \addplot[black,fill=cola1] table[y=small] {data/box/box.escape.txt};
            \addplot[black,fill=cola2] table[y=small_ablation] {data/box/box.escape.txt};
            \pvalue{1}{2}{1.25}{$<0.001$}
            \pvalue{3}{4}{1.25}{$<0.001$}
            \pvalue{5}{6}{1.25}{$<0.001$}
            
        \end{groupplot}
    \end{tikzpicture}
    \\
    \pgfplotslegendfromname{legendRQ2box}
    \caption{
        Distribution of the average velocity of escape for the best individuals found for each evolutionary run, obtained with three sizes and with or without pressure control.
        Dispensing with pressure control makes it impossible to escape from a cage.
        Numbers are $p$-values.
    }
    \label{fig:rq2-box}
\end{figure}
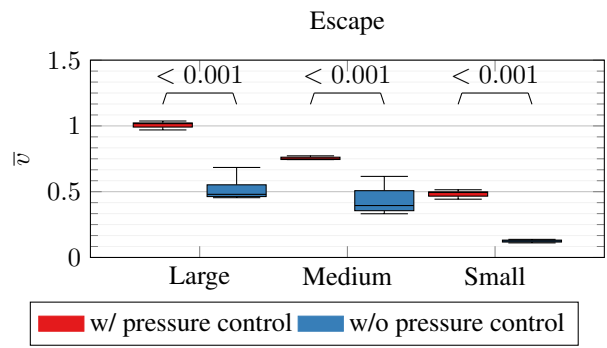

We summarize the results in \Cref{fig:rq2-evo}, which plots $\overline{v}$ in terms of median $\pm$ standard deviation for the best individuals over the course of evolution.
Moreover, \Cref{fig:rq2-box} reports boxplots for the distribution of $\overline{v}$ of the best individuals.
For every morphology, we also show the $p$-value for the statistical test against the null hypothesis of equality between the medians with and without pressure control.

\begin{figure}%[ht!]
    \centering
    \begin{tikzpicture}
        \begin{groupplot}[
            width=0.6\linewidth,
            height=0.5\linewidth,
            group style={
                group size=2 by 1,
                horizontal sep=1.5mm,
                vertical sep=1.5mm,
                xticklabels at=edge bottom,
                yticklabels at=edge left
            },
            every axis plot/.append style={thick},
            scaled x ticks = false,
            grid=both,
            grid style={line width=.1pt, draw=gray!10},
            major grid style={line width=.15pt, draw=gray!50},
            minor tick num=5,
            xlabel={Fitness evaluations},
            ymin=0,ymax=1.5
        ]
            \nextgroupplot[
                align=center,
                legend columns=3,
                legend entries={Large,Medium,Small},
                legend to name=legendRQ2evo,
                ylabel={$\overline{v}$},
                title={w/ pressure control}
            ]
            \linewitherror{data/line/evo.escape.txt}{i}{large_mu}{large_std}{cola1}
            \linewitherror{data/line/evo.escape.txt}{i}{medium_mu}{medium_std}{cola2}
            \linewitherror{data/line/evo.escape.txt}{i}{small_mu}{small_std}{cola3}
            
            \nextgroupplot[
                title={w/o pressure control}
            ]
            \linewitherror{data/line/evo.escape.ablation.txt}{i}{large_mu}{large_std}{cola1}
            \linewitherror{data/line/evo.escape.ablation.txt}{i}{medium_mu}{medium_std}{cola2}
            \linewitherror{data/line/evo.escape.ablation.txt}{i}{small_mu}{small_std}{cola3}
        \end{groupplot}
    \end{tikzpicture}
    \pgfplotslegendfromname{legendRQ2evo}
    \caption{
        Median $\pm$ standard deviation (solid line and shaded area) of the average velocity of escape for the best individuals found during each evolutionary run, obtained with three sizes and with or without pressure control.
        Our agent model is effective at shape-changing to escape from a cage.
    }
    \label{fig:rq2-evo}
\end{figure}
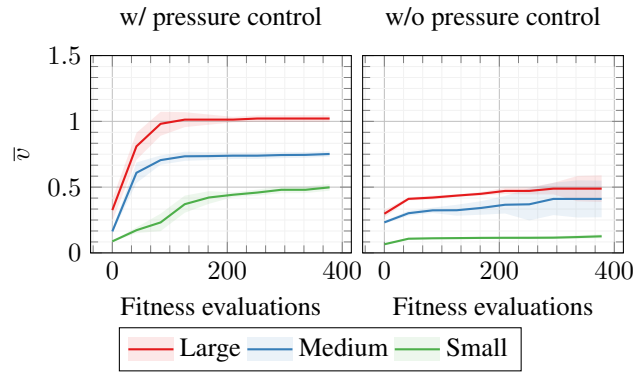

We remark that we say the task ``solved'' once all of the masses of a PSA are out of the cage; that happens when $\overline{v} \approx 1.0$ for Large, $\overline{v} \approx 0.75$ for Medium, and $\overline{v} \approx 0.5$ for Small.
From this consideration and the figures, we see that our proposed agent model effectively solves the task of escape from a cage.
We visually inspected the behaviors and found them to be highly adapted for an escape task.
Effective individuals decreased their internal pressure to reduce their area, almost flattening on the ground (see \Cref{fig:escaping} for a snapshot); then, they slithered through one of the apertures to successfully exit the cage (we remark that agents cannot evolve their initial pressure).
Albeit this turned out to be a recurring pattern, we observed some variations.
Some individuals, for example, evolved a repulsion for the walls: as soon as any of their touch sensors perceived a wall, they would contract themselves in the opposite direction.
The evolution of this trait might be due to the fact that, early in the optimization, we found many individuals to become tangled up as one of the walls wedged between two of their masses (joints, having no mass, cannot oppose to penetration).
Other individuals, when flattened, would literally crawl as big cats do when approaching preys, cautiously stretching out one mass after the other.

At the same time, evolution without pressure control did not find effective individuals.
From the right plot of \Cref{fig:rq2-evo}, we see that $\overline{v}$ barely departs from its initial value.
We visually inspected the evolved individuals, and found them to be not adapted at all for an escape task: all of them approached the walls to gain a little $\overline{v}$, but made no attempt at squeezing through the apertures.
Intuitively, the reason is their inability to control pressure, as they cannot shape change to effectively solve the task.
\Cref{fig:rq2-box} corroborates these findings by showing that $p$-values are significant for all three comparisons.

Through that evidence, we can answer positively to \ref{item:rq-2}: PSAs can effectively leverage shape change to solve a task that requires squeezing through a small aperture, after optimization.
Moreover, ablating the pressure control component of the controller results in no adaptation.
To the best of our knowledge, other works on soft robots solve this task by joint optimization of morphology and control \citep{zardini2021seeking,bhatia2021evolution}, which is complex, or morphology alone \citep{cheney2015evolving}, that might be less feasible than control alone in a real-world setting.
In the future, we envision such escape task to be the starting point of more interesting scenarios, like crawling inside caves with challenging terrain, as well as navigating ``claustrophobic'' mazes as cephalopods can do \citep{moriyama1997autonomous}.

\section{Conclusion}
\label{sec:conclusion}
Because of their infinite degrees of freedom, soft bodies pose unique challenges in terms of simulation, control, and optimization.
Here we propose a novel soft-bodied agents formalism, namely Pressure-based Soft Agents (PSAs): they are bodies of gas enveloped by a chain of springs and masses that simulates softness, with pressure pushing from inside the body and endowing the agent with structure.
Actuation takes place by changing the length of springs or modulating global pressure.
By virtue of such a mechanical model, PSAs can assume a large gamut of shapes.

We experimentally investigate whether it is possible to control PSAs and exploit their shape change potential.
That is what the paper demonstrates:
\begin{enumerate}[label=(\alph*)]
    \item we can control PSAs, as optimization finds effective controllers for a locomotion task on hilly terrain, a task that requires a decent degree of cognition to be solved;
    \item we can effectively exploit shape change for PSAs, as optimization finds effective controllers for the task of escape from a cage, a task that requires the agent to contort itself and squeeze through a small aperture. 
\end{enumerate}

Among the limitations of this work, it is worth mentioning that, having a low density, PSAs might not be suitable for object manipulation tasks.
At the same time, while we believe the model to be promising, as many real soft robots do rely on pressure to achieve shape change \citep{usevitch2020untethered,kriegman2021scale}, the manufacturability of PSAs is yet to be proven.
Future work will address these issues; for the moment, we agree with \citet{kriegman2019virtual} that virtual creatures can be ``as beautiful and complex as life itself''.
Indeed, we believe PSAs advance reality by providing a unified framework for soft-bodied agents that rely on shape change, so that several aspects can be tested prior to experimental implementation.
Other future directions include three-dimensional simulation, distributed controllers, the joint optimization of morphology and control (as already done for other soft agents \citep{medvet2021biodiversity,zardini2021seeking}), as well as the simulation of phenomena related to biological cells, such as phagocytosis and mitosis.
Opportunities are indeed many, and we have open-sourced our code with a gym \citep{brockman2016openai} interface to encourage usage by other researchers.

\section{Acknowledgements}
The authors wish to thank Eric Medvet for providing feedback on an early version of the manuscript.

\footnotesize
\bibliographystyle{apalike}
\bibliography{bib}

\end{document}

%% file: plot-macros.tex
\usepackage{tikz,pgfplots,pgfplotstable}
\usetikzlibrary{patterns}
\pgfplotsset{compat=newest}
\usepgfplotslibrary{groupplots,fillbetween,statistics}
\newcommand{\linesimple}[4]{
    \addplot [each nth point=1,#4,thick,no markers] table [x={#2},y={#3}] {#1};
}
\newcommand{\linewitherror}[5]{
    \addplot [each nth point=1,name path=minuserror,draw=none,no markers,forget plot] table [x={#2},y expr=\thisrow{#3}-\thisrow{#4}] {#1};
    \addplot [each nth point=1,name path=pluserror,draw=none,no markers,forget plot] table [x={#2},y expr=\thisrow{#3}+\thisrow{#4}] {#1};
    \addplot [forget plot,fill=#5,opacity=0.1] fill between[on layer={},of=pluserror and minuserror];
    \addplot [each nth point=1,#5,thick,no markers,legend image code/.code={\fill [fill=#5, opacity=0.1, draw=none] (0mm,-1ex) -- (0mm,1ex) -- (6mm,1ex) -- (6mm,-1ex) -- cycle; \draw [#5,thick] (0mm,0mm) -- (6mm,0mm);}] table [x={#2},y={#3}] {#1};
}

\newcommand{\pvalue}[4]{
    \draw ([xshift=.5mm]#1,#3) -- ([yshift=-1.5mm]#1,#3);
    \draw ([xshift=-.5mm]#2,#3) -- ([yshift=-1.5mm]#2,#3);
    \draw ([xshift=.5mm]#1,#3) -- node [above,scale=1] {#4} ([xshift=-.5mm]#2,#3);
}

\def\addlegendimage{\csname pgfplots@addlegendimage\endcsname}

\definecolor{cola1}{RGB}{228,26,28}
\definecolor{cola2}{RGB}{55,126,184}
\definecolor{cola3}{RGB}{77,175,74}
\definecolor{cola4}{RGB}{152,78,163}
\definecolor{cola5}{RGB}{255,127,0}
\definecolor{cola6}{RGB}{255,255,51}
\definecolor{cola7}{RGB}{166,86,40}

\definecolor{colb1}{RGB}{102,194,165}
\definecolor{colb2}{RGB}{252,141,98}
\definecolor{colb3}{RGB}{141,160,203}
\definecolor{colb4}{RGB}{231,138,195}

%% file: main.bbl
\begin{thebibliography}{}

\bibitem[Bhatia et~al., 2021]{bhatia2021evolution}
Bhatia, J., Jackson, H., Tian, Y., Xu, J., and Matusik, W. (2021).
\newblock Evolution gym: A large-scale benchmark for evolving soft robots.
\newblock {\em Advances in Neural Information Processing Systems}, 34.

\bibitem[Brockman et~al., 2016]{brockman2016openai}
Brockman, G., Cheung, V., Pettersson, L., Schneider, J., Schulman, J., Tang,
  J., and Zaremba, W. (2016).
\newblock Openai gym.
\newblock {\em arXiv preprint arXiv:1606.01540}.

\bibitem[Catto, 2011]{catto2011box2d}
Catto, E. (2011).
\newblock Box2d: A 2d physics engine for games.
\newblock {\em URL: http://www. box2d. org}.

\bibitem[Cengel et~al., 2011]{cengel2011thermodynamics}
Cengel, Y.~A., Boles, M.~A., and Kano{\u{g}}lu, M. (2011).
\newblock {\em Thermodynamics: an engineering approach}, volume~5.
\newblock McGraw-hill New York.

\bibitem[Cheney et~al., 2015]{cheney2015evolving}
Cheney, N., Bongard, J., and Lipson, H. (2015).
\newblock Evolving soft robots in tight spaces.
\newblock In {\em Proceedings of the 2015 annual conference on Genetic and
  Evolutionary Computation}, pages 935--942. ACM.

\bibitem[Cheney et~al., 2014]{cheney2014unshackling}
Cheney, N., MacCurdy, R., Clune, J., and Lipson, H. (2014).
\newblock Unshackling evolution: evolving soft robots with multiple materials
  and a powerful generative encoding.
\newblock {\em ACM SIGEVOlution}, 7(1):11--23.

\bibitem[Clapeyron, 1834]{clapeyron1834memoire}
Clapeyron, {\'E}. (1834).
\newblock M{\'e}moire sur la puissance motrice de la chaleur.
\newblock {\em Journal de l'{\'E}cole polytechnique}, 14:153--190.

\bibitem[Drotman et~al., 2021]{drotman2021electronics}
Drotman, D., Jadhav, S., Sharp, D., Chan, C., and Tolley, M.~T. (2021).
\newblock Electronics-free pneumatic circuits for controlling soft-legged
  robots.
\newblock {\em Science Robotics}, 6(51):eaay2627.

\bibitem[Ha, 2017]{ha2017evolving}
Ha, D. (2017).
\newblock Evolving stable strategies.
\newblock {\em blog.otoro.net}.

\bibitem[Hansen, 2006]{hansen2006cma}
Hansen, N. (2006).
\newblock {The CMA evolution strategy: a comparing review}.
\newblock In {\em {Towards a new evolutionary computation}}, pages 75--102.
  Springer.

\bibitem[Hansen, 2016]{hansen2016cma}
Hansen, N. (2016).
\newblock The cma evolution strategy: A tutorial.

\bibitem[Hansen et~al., 2019]{hansen2019cma}
Hansen, N., Akimoto, Y., and Baudis, P. (2019).
\newblock Cma-es/pycma on github.
\newblock {\em Zenodo, doi}, 10.

\bibitem[Hansen and Ostermeier, 2001]{hansen2001cmaes}
Hansen, N. and Ostermeier, A. (2001).
\newblock Completely derandomized self-adaptation in evolution strategies.
\newblock {\em Evolutionary Computation}, 9(2):159--195.

\bibitem[Hiller and Lipson, 2012]{hiller2012automatic}
Hiller, J. and Lipson, H. (2012).
\newblock Automatic design and manufacture of soft robots.
\newblock {\em IEEE Transactions on Robotics}, 28(2):457--466.

\bibitem[Hochner, 2012]{hochner2012embodied}
Hochner, B. (2012).
\newblock An embodied view of octopus neurobiology.
\newblock {\em Current biology}, 22(20):R887--R892.

\bibitem[Joachimczak et~al., 2016]{joachimczak2016metamrphosis}
Joachimczak, M., Suzuki, R., and Arita, T. (2016).
\newblock Artificial metamorphosis: Evolutionary design of transforming,
  soft-bodied robots.
\newblock {\em Artificial Life}, 22(3):271--298.
\newblock PMID: 27139940.

\bibitem[Kriegman, 2019]{kriegman2019virtual}
Kriegman, S. (2019).
\newblock Why virtual creatures matter.
\newblock {\em Nature Machine Intelligence}, 1(10):492--492.

\bibitem[Kriegman et~al., 2018]{kriegman2018morphological}
Kriegman, S., Cheney, N., and Bongard, J. (2018).
\newblock How morphological development can guide evolution.
\newblock {\em Scientific reports}, 8(1):13934.

\bibitem[Kriegman et~al., 2021]{kriegman2021scale}
Kriegman, S., Nasab, A.~M., Blackiston, D., Steele, H., Levin, M.,
  Kramer-Bottiglio, R., and Bongard, J. (2021).
\newblock Scale invariant robot behavior with fractals.
\newblock {\em arXiv preprint arXiv:2103.04876}.

\bibitem[Kriegman et~al., 2019]{kriegman2019automated}
Kriegman, S., Walker, S., Shah, D., Levin, M., Kramer-Bottiglio, R., and
  Bongard, J. (2019).
\newblock Automated shapeshifting for function recovery in damaged robots.
\newblock {\em arXiv preprint arXiv:1905.09264}.

\bibitem[Laschi et~al., 2016]{laschi2016soft}
Laschi, C., Mazzolai, B., and Cianchetti, M. (2016).
\newblock Soft robotics: Technologies and systems pushing the boundaries of
  robot abilities.
\newblock {\em Science Robotics}, 1(1):eaah3690.

\bibitem[Lipson et~al., 2016]{lipson2016difficulty}
Lipson, H., Sunspiral, V., Bongard, J., and Cheney, N. (2016).
\newblock On the difficulty of co-optimizing morphology and control in evolved
  virtual creatures.
\newblock In {\em Artificial Life Conference Proceedings 13}, pages 226--233.
  MIT Press.

\bibitem[Mast, 1911]{mast1911habits}
Mast, S.~O. (1911).
\newblock Habits and reactions of the ciliate, lacrymaria.
\newblock {\em Journal of Animal Behavior}, 1(4):229.

\bibitem[Matyka and Ollila, 2003]{matyka2003pressure}
Matyka, M. and Ollila, M. (2003).
\newblock Pressure model of soft body simulation.
\newblock In {\em The Annual SIGRAD Conference. Special Theme-Real-Time
  Simulations. Conference Proceedings from SIGRAD2003}, number 010, pages
  29--33. Citeseer.

\bibitem[Medvet et~al., 2020]{medvet20202d}
Medvet, E., Bartoli, A., De~Lorenzo, A., and Seriani, S. (2020).
\newblock {2D-VSR-Sim: A simulation tool for the optimization of 2-D
  voxel-based soft robots}.
\newblock {\em {SoftwareX}}, 12.

\bibitem[Medvet et~al., 2021]{medvet2021biodiversity}
Medvet, E., Bartoli, A., Pigozzi, F., and Rochelli, M. (2021).
\newblock {Biodiversity in evolved voxel-based soft robots}.
\newblock In {\em {Proceedings of the Genetic and Evolutionary Computation
  Conference}}, pages 129--137.

\bibitem[Moriyama and Gunji, 1997]{moriyama1997autonomous}
Moriyama, T. and Gunji, Y.-P. (1997).
\newblock Autonomous learning in maze solution by octopus.
\newblock {\em Ethology}, 103(6):499--513.

\bibitem[M{\"u}ller and Glasmachers, 2018]{muller2018challenges}
M{\"u}ller, N. and Glasmachers, T. (2018).
\newblock Challenges in high-dimensional reinforcement learning with evolution
  strategies.
\newblock In {\em International Conference on Parallel Problem Solving from
  Nature}, pages 411--423. Springer.

\bibitem[Nakajima et~al., 2015]{nakajima2015information}
Nakajima, K., Hauser, H., Li, T., and Pfeifer, R. (2015).
\newblock Information processing via physical soft body.
\newblock {\em Scientific reports}, 5(1):1--11.

\bibitem[Nolfi and Floreano, 2000]{nolfi2000evolutionary}
Nolfi, S. and Floreano, D. (2000).
\newblock {\em Evolutionary robotics: The biology, intelligence, and technology
  of self-organizing machines}.
\newblock MIT press.

\bibitem[Pfeifer and Bongard, 2006]{pfeifer2006body}
Pfeifer, R. and Bongard, J. (2006).
\newblock {\em How the body shapes the way we think: a new view of
  intelligence}.
\newblock MIT press.

\bibitem[Pigozzi, 2022]{pigozzi2022robots}
Pigozzi, F. (2022).
\newblock Robots: the century past and the century ahead.

\bibitem[Pritchard, 2001]{pritchard2001turgor}
Pritchard, J. (2001).
\newblock Turgor pressure.

\bibitem[Rieffel et~al., 2009]{rieffel2009automated}
Rieffel, J., Valero-Cuevas, F., and Lipson, H. (2009).
\newblock Automated discovery and optimization of large irregular tensegrity
  structures.
\newblock {\em Computers \& Structures}, 87(5-6):368--379.

\bibitem[Rus and Tolley, 2015]{rus2015design}
Rus, D. and Tolley, M.~T. (2015).
\newblock Design, fabrication and control of soft robots.
\newblock {\em Nature}, 521(7553):467.

\bibitem[Shah et~al., 2021a]{shah2021shape}
Shah, D., Yang, B., Kriegman, S., Levin, M., Bongard, J., and Kramer-Bottiglio,
  R. (2021a).
\newblock Shape changing robots: bioinspiration, simulation, and physical
  realization.
\newblock {\em Advanced Materials}, 33(19):2002882.

\bibitem[Shah et~al., 2021b]{shah2021soft}
Shah, D.~S., Powers, J.~P., Tilton, L.~G., Kriegman, S., Bongard, J., and
  Kramer-Bottiglio, R. (2021b).
\newblock A soft robot that adapts to environments through shape change.
\newblock {\em Nature Machine Intelligence}, 3(1):51--59.

\bibitem[Shepherd, 2006]{shepherd2006cytomatrix}
Shepherd, V.~A. (2006).
\newblock The cytomatrix as a cooperative system of macromolecular and water
  networks.
\newblock {\em Current topics in developmental biology}, 75:171--223.

\bibitem[Sims, 1994]{sims1994evolving}
Sims, K. (1994).
\newblock Evolving virtual creatures.
\newblock In {\em Proceedings of the 21st annual conference on Computer
  graphics and interactive techniques}, pages 15--22. ACM.

\bibitem[Singleton et~al., 2004]{singleton2004bacteria}
Singleton, P. et~al. (2004).
\newblock {\em Bacteria in biology, biotechnology and medicine.}
\newblock Number Ed. 6. John Wiley \& Sons.

\bibitem[Talamini et~al., 2019]{talamini2019evolutionary}
Talamini, J., Medvet, E., Bartoli, A., and De~Lorenzo, A. (2019).
\newblock Evolutionary synthesis of sensing controllers for voxel-based soft
  robots.
\newblock In {\em Artificial Life Conference Proceedings}, pages 574--581. MIT
  Press.

\bibitem[Usevitch et~al., 2020]{usevitch2020untethered}
Usevitch, N.~S., Hammond, Z.~M., Schwager, M., Okamura, A.~M., Hawkes, E.~W.,
  and Follmer, S. (2020).
\newblock An untethered isoperimetric soft robot.
\newblock {\em Science Robotics}, 5(40):eaaz0492.

\bibitem[Zappetti et~al., 2017]{zappetti2017bio}
Zappetti, D., Mintchev, S., Shintake, J., and Floreano, D. (2017).
\newblock Bio-inspired tensegrity soft modular robots.
\newblock In {\em Conference on Biomimetic and Biohybrid Systems}, pages
  497--508. Springer.

\bibitem[Zardini et~al., 2021]{zardini2021seeking}
Zardini, E., Zappetti, D., Zambrano, D., Iacca, G., and Floreano, D. (2021).
\newblock Seeking quality diversity in evolutionary co-design of morphology and
  control of soft tensegrity modular robots.
\newblock In {\em Proceedings of the Genetic and Evolutionary Computation
  Conference}, pages 189--197.

\end{thebibliography}
